# Exploiting Knowledge Graphs for Facilitating Product/Service Discovery

Sarika Jain

Department of Computer Applications, National Institute of Technology Kurukshetra, India
jasarika@nitkkr.ac.in

**Abstract**

Most of the existing techniques to product discovery rely on syntactic approaches, thus ignoring valuable and specific semantic information of the underlying standards during the process. The product data comes from different heterogeneous sources and formats giving rise to the problem of interoperability. Above all, due to the continuously increasing influx of data, the manual labeling is getting costlier. Integrating the descriptions of different products into a single representation requires organizing all the products across vendors in a single taxonomy. Practically relevant and quality product categorization standards are still limited in number; and that too in academic research projects where we can majorly see only prototypes as compared to industry.

This work presents a cost-effective solution for e-commerce on the Data Web by employing an unsupervised approach for data classification and exploiting the knowledge graphs for matching. The proposed architecture describes available products in web ontology language OWL and stores them in a triple store. User input specifications for certain products are matched against the available product categories to generate a knowledge graph. This mullti-phased top-down approach to develop and improve existing, if any, tailored product recommendations will be able to connect users with the exact product/service of their choice.

## 1 Introduction

The online shopping experience is evolving faster than ever before. The today's digitally empowered consumer wants seamless experience across all the touch points and expects retailers to be on the cutting edge of pricing and products. The companies are required to deliver a competitive price, a better checkout experience, immediate product availability, and above all, exact feature match of the product required. So many companies with lot many innovative products are purging up daily, increasing the information overload. This exponentially growing set of products confuses the customer with what to buy and from where to buy in order to quench the exact thirst and flavor of his requirement.

All the teaching and research institutes across the globe procure hardware infrastructure necessary for supporting the developmental and research activities of their projects. But due to poor understanding of the devices' technical specifications and characteristics, they face a lot of problems. With lot many variety of products available in the market, it becomes a daunting task understanding your options and then making the right choice. Consider simple queries like: "Why do I purchase a NAS when I already have file servers?", "What is the latest processor that I can afford for my new server?", OR "Should I go for Cloud Virtual Server?". There is a requirement of an interface between the customers and the retailers which can provide the best match between the triangulation of user's required specifications of the product, list of products available with those specifications, and the retailers selling those products; that could recommend to customer top n options to purchase from based on his provided specifications of the product.

Recommendation systems have found their usage in a multitude of applications including Information filtering, web personalization, and e-commerce. To list a few, companies like YouTube, Netflix, Spotify, Amazon, Pandora, and LinkedIn utilize the potential of recommender systems to facilitate product (videos, movies, music, products, friends, jobs) discovery. Various techniques for providing recommendations have been proposed in literature [Jain et al., 2015], most common being: the content-based, collaborative, and the hybrid approaches.

- In *content-based* (also called feature-based or reclusive) recommendation system, users have their preferences and items have their features; the system would correlate the target user's profile with the item's features in order to provide recommendations. Some example content-based recommendations include: "You are pursuing MCA201 course, the book XYZ is for you." and "You are a marketing guy; this new smart phone is for you".
- *Collaborative filtering* systems make recommendations based on historic users' preference for items. They utilize some similarity algorithms to predict best match either based on similar users' preferences or based on the ratings of similar items [Sehgal et al., 2016]. Some example collaborative recommendations include: "You generally buy tomato soups, here is a new flavor of tomato soup you will definitely like", "You generally buy tomato soups, here is the most popular flavor of the same", "Your friend X has bought this", and "You have just purchased paint for your house. People who bought paint also purchased a painting".
- *Hybrid approaches* work upon a unifying model incorporating information from both users/items' characteristics and user-item interactions, i.e., both the similarity of product attributes and the purchase history. An example of hybrid, i.e., content-based collaborative recommendation is "You just bought that book. This one is also from the same author".

All known approaches fail in scenarios where (a) new innovative products are introduced, (b) existing products disappear from the catalogue, (c) custom made products where each product is unique, (d) the customer has very less knowledge of the specifications. Current research focuses on the application of deep learning, exploitation of knowledge graph, reinforcement learning, and user profiling to provide efficient explainable recommendations along with providing user privacy. The choice of strategy to be adopted much depends on the strategic goal under consideration. This work proposes a multi-phased top-down approach to develop and improve existing, if any, tailored product recommendations to connect users with the exact product/service of their choice. We exploit the Semantic Intelligence Technologies (SITs) in order to suggest for new products in the market based on the user's input features. We choose the engineering equipment as the domain and provide an engineering equipment recommender (EER) as an outcome of this work. The EER comprises of following major phases, the creation of product ontology, the ingestion and integration of data from product catalogues; product classification, the storage of data in triple store; and exploring, search and consumption of stored knowledge.

## 2  Overview and Proposal

Nothing comes without challenges. Making the best match between the offers (product catalogues), the retailers making offers, the requested specifications, the customer's preferences, and the intended use of the product comes as the basic problem in product discovery. While looking for the best match in product offerings, the customer faces following problems:

1. *Uncertainty (Incomplete Specifications):* On one hand, the buyers are unaware of their specific requirement. They have a vague and very general idea of what they wish to purchase. On the other hand, the seller may miss some important descriptions while listing the items in his product catalogue. This problem of incomplete specifications require to be flexible enough and adjust whenever constraints changes. The uncertainty due to incomplete representations may result in duplicate entries in product knowledge base being created, and makes product matching difficult due to data sparsity on both sides (buyer and seller).
2. *Monopoly:* Monopoly comes as a challenge in e-commerce when it is the question of rare and niche products, i.e., custom made products where each product is unique; like leather iPad cases, a unique beaded necklace, and antiques. These are also termed as one-of-a-kind or on-demand products. These types of products are sold by a single vendor and cannot be cross-validated for their category while populating the product knowledge base.

3. ***Scalability (New Product Offers and Schemes):*** When new innovative product offers are introduced, it becomes a challenge to treat new products and categorize them (specifically when they are bringing in new attributes). Schemes such as bundled products and lots of similar products also add a challenge to product categorization. Many different products can be bundled in many possible manners; also similar products can be clubbed in different numbers to make different lots.
   - This increases the number of classes multi-fold leaving the solution unscalable.
   - It requires retraining the model each time an unseen product offer or scheme is introduced.
4. ***Products soon Become Outdated:*** When existing product offers disappear from the catalogue, or the schemes offered are no more alive, it again requires retraining the model.
5. ***Heterogeneity (Non-Standardized Catalogues):*** Another major challenge is the global hierarchy of product classes. The product catalogues are not standardized; different catalogues have different descriptions and representations for the same product class. The effective exploitation of the product catalogues requires finding relevant data sources and integrating them by combining the elements.
6. ***Natural Language Queries:*** The user queries may be unambiguously specified. For e.g. "Suggest a laptop with maximum possible memory". It is not clear how much memory is sufficient and whether RAM is queried or Hard Disk.

## 2.1 Previous Key Contributions

Knowledge graphs have tremendous potential to exploit the data of web stores for better recommendations. Previous research has depicted the importance of SITs in information systems.

**Domain Vocabularies**

Pohorec et al. (2013) analyzed and compared different approaches for semantic tagging of machine-readable data on the web. The authors discuss formats available: RDFa, Microdata and Microformats. Hepp, in 2005 proposed an ontology vocabulary eClassOWL to enable the representation of catalogues. Brunner et al. in 2007 provided an overview of the improvements that are possible for managing product data. However at that time SITs were not mature enough and lacked standards, Brunner presented an information system built on semantic technologies for discovering product data. Then in 2008, Hepp presented another ontology GoodRelations for data representation of products. The Aletheia architecture proposed in 2010 federates information from heterogeneous sources [Wauer, 2010]. Fitzpatrick et al. in 2012 presented a holistic approach for master data management. This work focused on the usage of the same data structures in every phase of the product life cycle. In 2013, Stolz et al. presented one more approach focusing on integration. The authors proposed transformation from the BMECat catalogue format to GoodRelations ontology.

Nederstigt et al. (2014) proposes FLOPPIES for semi-automatic population of ontology from Web stores. Vandic et al. (2012) is an example of the use of semantic tagging products in Internet stores. Platform as a source uses its own database that is constantly updated by pinging web shops that use RDFa for semantic labeling. Recent works [Meusel et al., 2015] have shown Microdata format as the most commonly used markup format, along with highest domain coverage. They have also commented that schema.org vocabulary is most frequently used to describe products.

**Classification**

Mai et al., 2018 compared the title-based and full text-based classifiers over the scientific literature and found that the title-based classifiers can outperform if provided with sufficient training data. GoldenBullet [Ding et al, 2002] compares the accuracy of Vector Space Model, kNN classifier, and Naïve Bayes over a set of manually pre-classified product data and concluded that Naïve Bayes outperforms. Kim et al, (2006) provided a mechanism of utilizing the structure in the data for improving the performance of Naïve Bayes classifier. Abbott et al. (2011) in collaboration with NewZealand based company Unimarket proposed a proof-of-concept kNN classifier.

The supervised approaches to classification demand for huge datasets for training the classifier. The WebDataCommons project has produced a training dataset for evaluating Large-Scale Product Matching methods [Primpeli et al., 2019]. Different approaches have been utilized to go away with the demand of lots of data for supervision. Ristoski et al. (2016) exploits the convolutional neural network and deep learning techniques for product matching and categorization for unstructured product descriptions. Decker et al. (2007) and CSO Classifier [Salatino et al., 2019] utilizes keywords and abstract of articles as metadata and performed unsupervised classification over computer science topics.

**Scope**

Most of the existing techniques to product classification and matching rely on syntactic approaches, thus ignoring valuable semantic information during the process. Despite all the efforts, we still lack practical unsupervised approaches for classifying entities according to a granular set of categories. The existing works do not properly reflect the specific semantics of the underlying standards. The number of quality, practically relevant product ontologies on the Web is still limited [Roos et al. 2002], because most ontology engineering work is done in the context of academic research projects where efforts rarely go beyond toy status. Thus, a cost-efficient solution able to accommodate business needs on the Data Web would be greatly appreciated. This work proposes the inclusion of knowledge bases information in order to improve classifications tasks of products using their metadata. The direction towards learning automatic semantic relationships on unstructured sequence of e-commerce text has been of less focus. Integrating the descriptions of different products into a single representation requires organizing all the products across vendors in a single taxonomy.

## 2.2 System Architecture and Framework

In order to overcome the challenges of the e-Commerce marketplace, this work aims at laying the foundation of a search and price comparison semantic web portal for product offers provided by various eShops. The idea is to provide information and tools to help the customer find the best product across various eShops, in the same manner as trivago suggests the best hotel deal according to the customer requirements. The portal provides two interfaces: one to the new products from manufacturers/retailers/eShops to enter into the knowledge base, and the other to the end users to get suggestions on which products to purchase. The end user can input raw technical specifications of the product to be purchased and gets the top n recommendations (in descending order of similarity) on the products available in eShops. Each recommended product is accompanied with the detailed specifications including the OEM, retailer, cost and the overall similarity score with the input specifications.

What is required is a content management system with semantic intelligence that is more than just a repository of unstructured documents, or data stored in a SQL database, or even NoSQL databases. The most pragmatic approach is to look at the strengths and capabilities that are existent in our own data and leverage them by practices like metadata, taxonomies, ontologies, and knowledge graphs. The recommendation engine will be powered by the offbeat semantic intelligence technologies and rich meta data to replace the manual process and allow for advanced AI capabilities and high quality and targeted recommendations with explanations to why the said recommendation.

Figure 1 presents an architecture proposed for facilitating recommendations utilizing the web semantics. The proposed system architecture comprises of majorly two phases / modules and the underlying knowledge base. The knowledge base is an ontology-based representation of domain concepts with all the offers and schemes available in the TBOX and all the known individuals available in the ABOX.

**Phase 1 – Knowledge Modeling –**

The system admin generates a taxonomy and ontology that is comprehensive, sophisticated, user friendly, and applicable to multiple use cases. The content, i.e, the product instances are extracted from different sources (structured or unstructured) and knowledge graph is created in accordance with the developed ontology.

**Phase 2 – Knowledge Consumption –**

The ontology as constructed in phase 1 is useful for various use cases as resource manager, semantic search, knowledge analytics, question answering, and more.

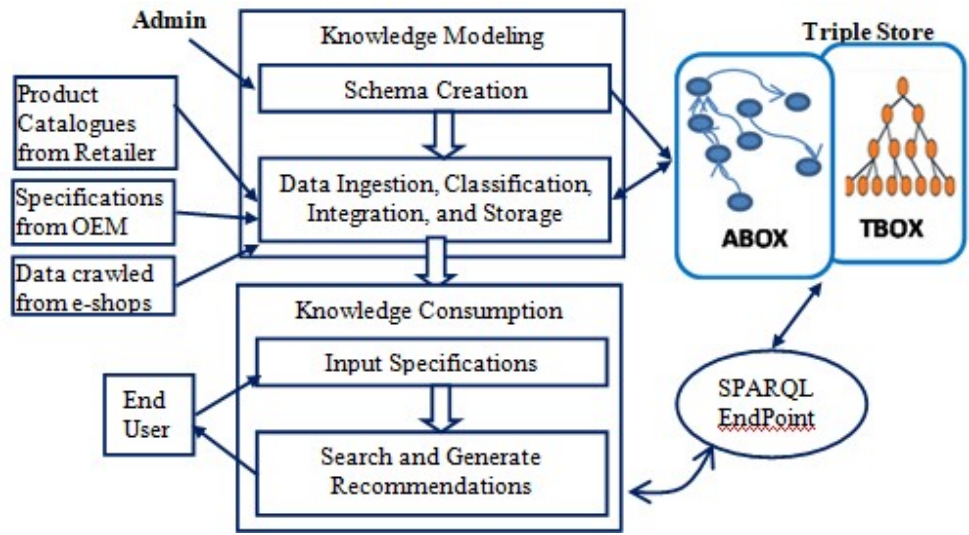

**Fig 1. System Architecture**

## 2.3 Benefits of the Approach

Graph is a flexible and self-explaining structure. A knowledge graph has been used to present (parts of) the ontology which have been selected by the classifier. The proposed approach facilitates operations over aggregated data toward automating the tasks of e-business. It assists in smart analytics by providing automatic indexing and semantically enhancing the meta data of engineering equipment. EER is capable of detecting trends in design of engineering equipment for better understanding. The system will be made available through a semantic web portal with following specific benefits:

a) *Rapid Information Retrieval and Search:* Data and information can be navigated, rapidly and efficiently in hierarchical, textual and graphical forms. A better search and discovery of products/services is provided across the vendors. It enables users to explore the product catalogues and provide feedback at different levels.
b) **Smart and Automatic Classifier:** Automatically annotates engineering equipment with the topics available in EEO.
c) **Smart Recommender**: This smart recommender recommends a list of equipment available based on the input specifications. For every product or service required, a lot of products and services are available in the market and too provided by a number of vendors. It becomes difficult for a customer to select which product/service to purchase and from which vendor. It helps the manufacturer also to select equipment to market at specified avenues. If the requirement of the user (/avenue) is known in the form of specifications, the system can generate recommendations on which equipment best fits the requirement specification.
d) *Supports more reliable, less ambiguous, communication:* Disparate companies and different groups of people use different nomenclature for the same thing. By establishing a common technical language, eShops as well as end users will be able to process and analyze data automatically.
e) *Avoid Duplication of Work:* Categorizing the things globally and storing at a common place promises reuse of specifications leading to cost reductions.
f) *Relational Synergy:* The related data is now linked, hence providing synergy. The user comes to know additional related information that was not possible in traditional approach.

## 3 Materials and Methods

This work motivates the utilized the most robust set of tools and techniques that makes extensive use of AI.

### 3.1 Why Semantic Intelligence

Data is the heart and core for all AI tasks. The focus is on data as compared to process. More importantly, quality and optimized data enables better understanding and information sharing for better decision making [Patel and Jain, 2019]. Semantic Intelligence refers to filling the semantic gap between the understanding of humans and machines by making a machine look at everything in terms of object oriented concepts as a human look at it. Semantic Intelligence Technologies (SITs) is the amalgamation of core AI features such as ML and NLP, and semantic web technologies such as RFD, OWL, SPARQL etc. Intelligence Technologies are capable enough and offer ways to not only handle the original 3 V's of Big Data (Volume, Velocity and Variety); but also the additional 3 V's (Veracity, Variability and Value).

### 3.2 Why Meta Data, Taxonomies and Ontologies

To provide recommendations over the input specifications, the product information (either in catalogues or on web) from the retailer need to be machine understandable, hence require to be annotated semantically. Different forms of metadata and pre-defined vocabularies are useful for semantic annotation of content. The information that expresses information about the meaning and context of some data is ***metadata.*** Metadata helps in fetching all related content about a data precisely. Because of the limited support by people creating websites and a very slow and non-existent adoption of semantic technologies, some technologies started combining semantic information along with the already existing content by increasingly adopting Microformats, RDFa, and Microdata as semantic markup languages [Meusel et al., 2014].

These approaches introduce ambiguity as the same thing can be annotated in different ways by different people. To normalize the terms for providing good metadata, a pre-existing taxonomy is required. Either use some existing taxonomy, or author one. After authoring taxonomy, it is to be published on LOD cloud.

***Taxonomies and Thesauri*** are very closely related controlled vocabularies to organize relevant data by describing relations between concepts and their labels including synonyms. They allow for effective classification and categorization of both structured and unstructured information for the purposes of findability and discoverability.

***Ontologies*** are generalized semantic data models that define the vocabulary of a domain and the relationships between the concepts of vocabulary. Ontologies standardize the terminology of an application domain, and present a unified view of data sources, hence making information sharing and data integration easy.

Metadata, Taxonomies, and ontologies together bring structure to the unstructured data and provide information to the machines in a machine-readable format, so that meaningful inference can be drawn.

We have decided to leverage ontologies for knowledge graphs schema

### 3.3 Why Knowledge Graph

Many organizations have begun bringing Knowledge Graphs in their information ecosystem as they come with such technologies and tools that solve most of the problems they face, like interoperability, findability and more [Curry and Adeboyega, 2020; Li et al., 2020]. Knowledge graphs are very close to the human thinking as they are built upon RDF data model, thereby allowing us to capture relationships in the same manner as the human brain processes information. The knowledge graphs organize large amounts of data and information. A knowledge graph (KG) is a very large semantic network that represents knowledge without a strict schema by integrating information from variegated heterogeneous sources. Examples include Wordnet, Google knowledge graph, DBPedia, Yago, WikiData. No clear definition of knowledge graph exists till date and the term is fundamentally used interchangeably with ontology. The most general view calls the complete thing (concepts, relationships, and instances) a knowledge graph that can be queried for instances, and the schema only (i.e., concepts and relationships) an ontology that can be used for inference. Other views exist with disagreements on size, and role of semantics. We will take the general view here and speak of the ontology as the schema / data model and the KG as the actual instance data (like rows in a RDBMS) built in accordance with some ontology (schema).

The classes and relationships are termed as TBox (Terminological Box) and the instances termed as ABox (Assertional Box) in Description Logic; both of them together making up a knowledge graph. TBox statements are used and stored as a schema, while ABox statements are stored as instance data within transactional DBMSs. The NOSQL databases and the RDF databases, TBox statements (data model) and ABox statements (data/instances) are stored using the same approach. In view of all above, KG = ONTOLOGY (TBox) + DATA (ABox). In a knowledge graph, the TBox is not so big but they have a very large ABox. The readers have to make a note that every RDF graph (like the marks in course Semantic Web of students of a university) is not a knowledge graph as it may not capture the required semantics.

## 3.4 Product Categorization

Lots of studies have been focused on structuring the product catalogs, i.e., characterizing things (products, articles, human beings etc.) with categories (Mauge et al., 2012). The task has traditionally been addressed in two ways.
1. Generate Topics from Scratch: This approach generates categories from the text in a bottom-up style using clustering [Bolelli et al., 2009; Griffiths et al., 2004]. One of the first works in this approach to identify categories is the Topic Detection and Tracking (TDT) program developed by DARPA [Allan et al., 2998] in the domain of broadcasted news. To organize a stream of documents, the TDT program performs cluster analysis. Some methods use keywords and phrases and perform analysis to identify clusters [Duvvuru et al., 2013; Wu et al. 2017]. Yet some other methods utilize word embeddings to quantify semantic similarities in unstructured text [Zhang et al., 2018].
2. Leveraging a domain vocabulary or ontology: This approach uses classifiers that assign a set of pre-existent categories (like schema.org) to the given product description with an advantage to annotate the thing with clean and formally-defined categories.

The first set of approaches tends to produce noisier and less interpretable results [Osborne and Motta, 2012]. The latter solution relies on a set of well defined product categories, but requires a good vocabulary in the domain. Ontologies are fundamental to ensure the consistency between datasets since they formally represent and precisely define concepts and their relationships. They improve the semantic content of the data and link datasets at a schema level. This second set of approaches may utilize supervised machine learning, deep learning, or unsupervised machine learning.

## 3.5 Categorization Standards

We need to create the schema, i.e., the TBox first. Various categorization standards exist for products and services, the major among them being
- the United Nations Standard Products and Services Code (UNSPSC) standard is available in 14+ languages and is currently managed by GS1 US,
- GS1's Global Product's Classification (GPC) offers a universal set of standards for every type of product.
- eOTD is the ECCMA Open Technical Dictionary of unambiguously described concepts
- European Union's Common Procurement Vocabulary (CPV) is a vocabulary of services to process the procurement and tendering.
- eCl@ss OWL is the OWL ontology for products and services on the semantic web.
- Schema.org: Google, Yahoo, Yandex, and Microsoft are the four major players who have created and are maintaining schema.org as a vocabulary for structured data on the web. This vocabulary can be used with RDFa, microdata, and also JSON-LD. Using Schema.org on meta-modeling level enables integration to the semantic web vocabulary used by many search engines (e.g. Google) and e-commerce sites (e.g. eBay).
- Good Relations ontology, an OWL Lite ontology derived from the categorization standard eCl@ss 5.1 that covers the representational needs of typical business scenarios in the commodity segment.
- Product Ontology: The product structuring knowledge can be represented in the form of a formal ontology expressed in Web Ontology Language (OWL). Semantic Web Rule Language (SWRL) can extend the OWL for providing capabilities to express constraints that are not expressible as OWL axioms. All product configuration constraints can be expressed as SWRL rules making the solution similar to rule based approaches, allowing to reason on the ontology using inference engines. One such ontology is Product ontology that extends Schema.org and Good Relations vocabularies with links to hundreds of thousands precise definitions for types of product in Wikipedia.

## 3.6 Triple Store

Because of the structure of knowledge graphs, they should be backed by a RDF data store in order to exploit the embedded context. Triple stores serve the road ahead from Big Data to Smart Data. For triple stores, there is a choice between triple stores built from ground up, i.e. RDF triples (like Jena TDB, GraphDB), also called native triple stores; and the triple stores that are based on relational databases (like Virtuoso, Jena SDB) and simply provide a SPARQL interface, also called non-native triple stores. We have chosen the first option as the internal RDF graph structure can handle the scale, speed, and variety of data and is flexible enough not to need its schema re-defined every time new data is added.

Triple stores have lot of foreseen benefits for storing RDF data. They are queried over existing HTTP, so no bridging solutions are required. They allow multiple values for a variable thus facilitating multiple internationalized languages. RDF is more expressive than any other language to model complex and unstructured data. As all the triple stores speak the same language, moving data between stores is easier.

# 4 Knowledge Modeling

The knowledge of product offers is to be represented in the form of semantically enriched data and then hosted. To describe the data of product catalogues, metadata is added by using a self-curated vocabulary.

## 4.1 Schema Creation

In the knowledge graph building efforts, the first question that arises is the choice of vocabulary to use. We have built an ontology-based representation of the product catalogue in a manner that it is interoperable and can be combined with offers from other retailers. We call it the engineering equipment ontology (EEO). In this knowledge modeling task, we formalize the catalogue by representing the product descriptions as ontological classes in Web Ontology Language (OWL) and store them in a triple store. The EEO will serve as the meta-knowledge for the products and/or services and will be updated periodically for any new description or product type that comes in.

We will identify our classes (Product, Laptop, Desktop …), relationships, and attributes. The relationships and attributes both can be considered as properties. Relationship between two classes is termed as object property, and attribute of a class is termed as data property. Some properties of the Product class might be:

Product → has price → Some Number
Product → has name → Some String
Product → has description → Some Text
Product → has price currency → Currency
Product → has make → OEM
Product → is available with → Retailer
Product → has category → Some subclass of Product

Here we list some other properties of other classes.
Laptop → is a → Product
Retailer → located in → Location
OEM → based in → Location
Because of open world assumption (OWA) in ontologies, we capture possible properties that could apply to many, but not necessarily all products. The EEO ontology will serve as a general data model and a reusable framework, we will use to describe the instances as we discover them. Figure 2 depicts the graphical view of EEO.

## 4.2 Data Ingestion

When the graph model, EEO is ready, we start applying it to the incoming data (product catalogues), and in effect populate the graph with data, i.e., instances of product offers. There are two types of data sources, structured and unstructured. In cases when the data source is structured like a relational database, or the data is highly dynamic; we prefer to keep the data in-place and utilize virtual graphs as a mapping from SPARQL to SQL. This strategy maintains the freshness of data as it is always live and avoids any duplication, but degraded performance while querying. In cases when the data source is unstructured, or less dynamic; we should utilize the ETL (Extract, Transform, Load) to extract data from the source periodically, transform it into RDF, perform semantic tagging using ontology, and load into the graph. As product catalogues are highly unstructured and mostly static about product offers, except the newly introduced products, we take the ETL approach.

***EEO Classifier:*** Better classification increases the discoverability of products. We assign predefined category labels to product instances (eg. Macbook Air is a Laptop and also Product). Since the EEO is in its infancy and has not yet been published on LOD (Linked Open Data) cloud for wide and routine use by researchers yet, adopting the supervised machine learning algorithms is not possible as it would require a sufficiently good number of training examples for all the relevant concepts/categories. We utilize a new unsupervised approach that automatically classifies equipment according to the categories in EEO by measuring similarity. The Equipment Classifier works by taking the metadata associated with equipment (type, features, and commercial details) as input and returns a selection of categories level-wise drawn from the equipment ontology. This classifier will allow all the relevant stakeholders to annotate their equipment data according to EEO.

We assume that the product catalogues consists of one product per page. First, all stop words are removed from the product catalogue, and the resulting list of words for each product is stemmed. Here we extract key terms from the product catalogues by the use of custom Named Entity Recognition (NER) models. This phase utilizes the power of NLP to rapidly harvest information from the thousands of products in the catalogues. After the preprocessing phase, we have a list of product descriptions in a JSON file with one product instance per line using the following schema {productID, name, description, price, priceCurrency, make, category}. Not only the product catalogues, but the data from the web can be crawled and converted into this format for further classification.

***Knowledge Graph Creation:*** A knowledge graph is a network structure that captures the rich knowledge contained by items. It provides abundant reference values for providing recommendations, as it expands the amount of information of each item and strengthens the connections between them. Using our ontology as a framework, we can add in real data about individuals, say "Macbook Pro", "Lenovo Thinkpad", "Dell Precision 7920 Tower", and "Apple Mac mini" to create an Engineering Equipment Knowledge Graph (EEKG). With the information in Product catalogues, as well as our ontology, we can create specific instances of each of our ontological relationships. For example, if we add in all of the individual information that we have about one of our products, Macbook Pro, we can see the knowledge graph as in figure 3. Using this knowledge graph, we can view our data as a web of relationships, instead of as separate tables, drawing new connections between data points that we would otherwise be unable to understand by the relational tables.

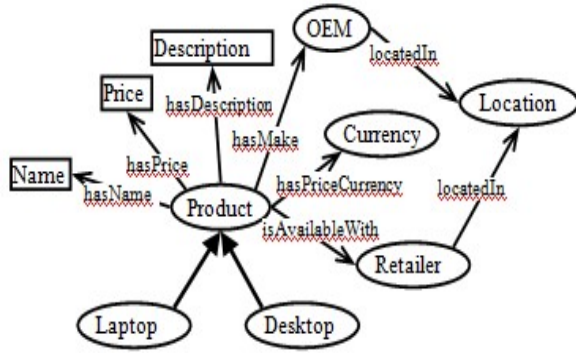 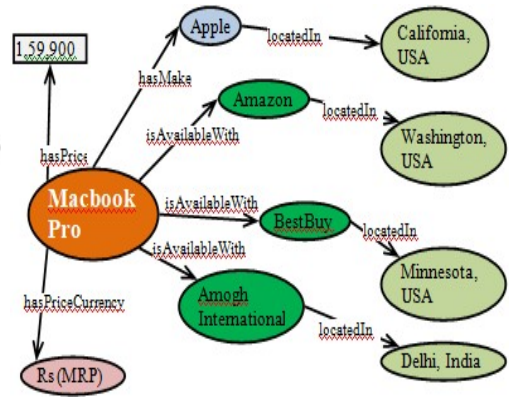

**Fig 2. Graphical View of EEO**     **Fig 3. Engineering Equipment Knowledge Graph (EEKG)**

## 4.3 Knowledge Hosting

Two different approaches are utilized for storing this semantically annotated data. If this data is meant to be exploited by some use case in a manner to be understandable by the automated agents, for say SEO then it is stored in a JSON-based document database, e.g. MongoDB. In this case, the data is queried over an API and no RDF reasoning is possible. If the data is meant to be exploited by the enterprises for reasoning, it is to be stored in a triple store and querying is simplified through SPARQL endpoint.

As we have to use our data in our site itself for recommendation, we take the second approach of storing in triple store. The RDF triple stores are based on W3C standards and they store triples (subject-predicate-object). Amongst the open source and native implementation of triple stores, we had a choice between Jena TDB, OpenLink Virtuoso, Blazegraph, RDFox, Sesame (rdf4j), Stardog. We have chosen the Jena TDB by Apache Foundation.

## 5  Knowledge Consumption

Now all the required knowledge is available in a machine understandable format in the triple store, it is time to query it and fetch valuable insight for various use cases. We talk about generating recommendations of top n matched offers in the triple store. The representation describes products in terms of class descriptions in the web ontology language OWL. User requests for certain products are also represented as OWL concepts and matched against the available product categories to generate a knowledge graph. This process of categorization of input specifications iterates over a period of time by suggesting the user to refine the specifications and in turn rearranging the knowledge graph. After the knowledge graph is in place, it is matched with the available product instances and top n product offers are displayed to the user along with the attributes.

1.  User states the keywords of required specification.
2.  SPARQL query is generated with the input specifications
3.  Product instances corresponding to matched specifications are listed as output.

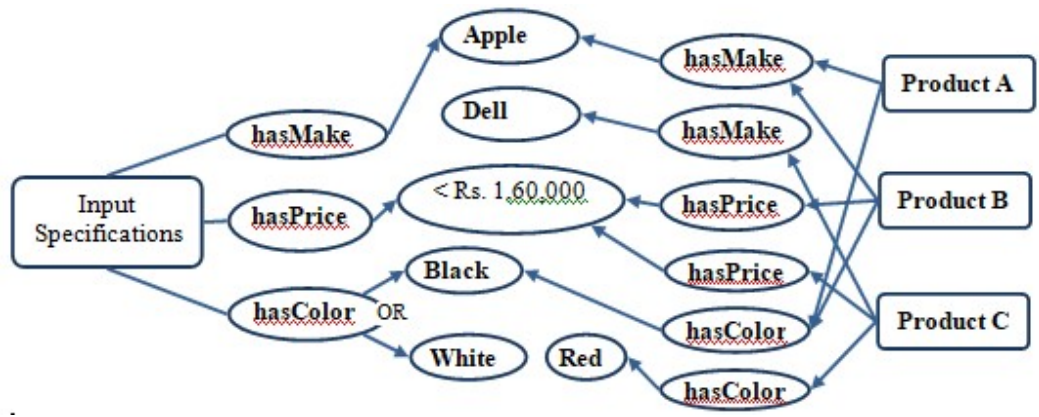

Fig 4. Product Matching

Product B is the best choice with make Apple, price less than Rs. 1,60,000 and color Black as desired. Product matches two features, and product C matches only one.

## 6 Findings and Conclusion

The product descriptions on semantic web are inherently complex and in heterogeneous formats. This work has provided a common comprehensive platform for representing and storing different product descriptions. It has utilized an unsupervised approach to classification with no need of labeled dataset for training; and exploited the knowledge graphs for understanding far-flung attributes of products to calculate similarity with the input specifications. As an outlook into the future, we envision to design and evaluate web ontology for products as a compatible extension of schema.org along with comparing our approach with state-of-the-art vocabularies and recommendation systems for e-commerce.